\documentclass[11pt]{article}

\usepackage[final]{acl}

\usepackage{times}
\usepackage{latexsym}
\usepackage{url}
\usepackage{amsmath,amssymb}
\usepackage[ruled,vlined]{algorithm2e}
\usepackage{xcolor}
\usepackage{booktabs}
\usepackage{multirow}
\usepackage{longtable}
\usepackage[most]{tcolorbox}

\newtcolorbox{promptbox}{
  colback=gray!10,
  colframe=gray!60,
  arc=4mm,
  boxrule=0.5pt,
  width=\linewidth,
  boxsep=8pt
}

\usepackage[T1]{fontenc}

\usepackage[utf8]{inputenc}

\usepackage{microtype}

\usepackage{inconsolata}

\usepackage{graphicx}

\title{Letting Tutor Personas Speak Up for LLMs: Learning Steering Vectors from Dialogue via Preference Optimization}

\author{
 \textbf{Jaewook Lee\textsuperscript{1}},
 \textbf{Alexander Scarlatos\textsuperscript{1}},
 \textbf{Simon Woodhead\textsuperscript{2}},
 \textbf{Andrew Lan\textsuperscript{1}}
\\
 \textsuperscript{1}University of Massachusetts Amherst,
 \textsuperscript{2}Eedi
\\
 \small{
   \texttt{\{jaewooklee,ajscarlatos,andrewlan\}@cs.umass.edu}
   }, \small{
   \texttt{simon.woodhead@eedi.co.uk}
 }
}

\begin{document}
\maketitle
\begin{abstract}
With the emergence of large language models (LLMs) as a powerful class of generative artificial intelligence (AI), their use in tutoring has become increasingly prominent. Prior works on LLM-based tutoring typically learn a single tutor policy and do not capture the diversity of tutoring styles. In real-world tutor–student interactions, pedagogical intent is realized through adaptive instructional strategies, with tutors varying the level of scaffolding, instructional directiveness, feedback, and affective support in response to learners’ needs. These differences can all impact dialogue dynamics and student engagement. In this paper, we explore how tutor personas embedded in human tutor-student dialogues can be used to guide LLM behavior without relying on explicitly prompted instructions. We train a steering vector using preference optimization: an activation-space direction that guides model responses toward specific tutor personas. We find that this steering vector captures tutor-specific variation across dialogue contexts, improving semantic alignment with ground-truth tutor utterances and increasing preference-based evaluations, while largely preserving lexical similarity. Analysis of the learned scaling coefficients further reveals interpretable structure across tutors, corresponding to consistent differences in tutoring behavior. These results demonstrate that activation steering offers an effective and interpretable way for controlling tutor-specific variation in LLMs using signals derived directly from human dialogue data.
\end{abstract}

\section{Introduction}
With the emergence of large language models (LLMs) as a powerful class of generative artificial intelligence (AI), their use in educational tutoring has become increasingly prominent, spanning both general-purpose systems~\cite{gemini,chatgpt} and education-specific tutors~\cite{livehint,khanmigo}. 
%
%
Recent work has also connected LLMs with Knowledge Tracing (KT), using KT-derived or LLM-assisted student-state estimates for tasks such as student error simulation and mastery modeling in tutoring dialogues~\cite{huang2026interpretable,duan2026kaser,scarlatos2025exploring}.

Approaches to building LLM tutors vary in how pedagogical behavior is specified and optimized. Many approaches simply prompt LLMs with dialogue policies and guardrails to elicit best-practice pedagogy, including reducing solution leakage, encouraging scaffolding, and enforcing Socratic-style questioning~\cite{pal2024autotutor,puech2025towards,zhang2024spl}. Other approaches instead train LLMs using supervised fine-tuning (SFT) to mimic human tutor behavior or use reinforcement learning (RL) on synthetic tutoring dialogues. These approaches optimize proxies of pedagogical quality, such as improving feedback correctness and alignment with tutoring goals~\cite{scarlatos2024improving}, directly optimizing for pedagogical alignment signals derived from preference data over tutoring behaviors~\cite{sonkar2024pedagogical}, or leveraging pedagogical rubrics and LLM-based simulated students in-the-loop to encourage greater learning outcomes~\cite{dinucu2025problem,scarlatos2025training}.

\begin{figure*}[t]
    \centering
    \includegraphics[width=1\linewidth]{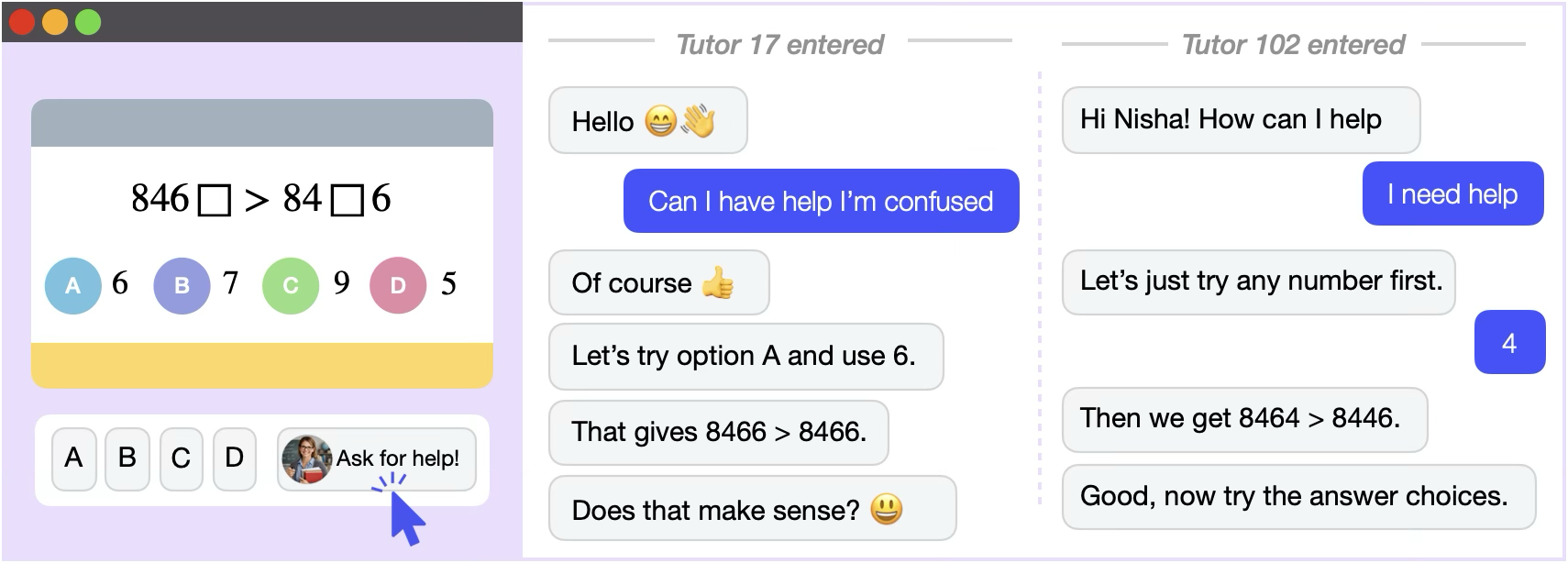}
    \caption{Real math tutoring dialogue on inequalities~\cite{eedi-dialogues}, showing how tutors 17 and 102 differ in instructional style when solving the same problem.}
    \label{fig:examples}
\end{figure*}

Despite making improvements on pedagogical quality,
these approaches only learn a single tutor policy and do not attempt to capture the diversity of tutoring styles.
Figure~\ref{fig:examples} illustrates how real tutors differ in their instructional style when addressing the same problem. One noticeable difference is the use of emojis and affective language, which conveys emotional support and rapport. Another key difference is instructional approach: the first tutor provides increased emotional support and guides the student step-by-step, whereas the other provides minimal rapport-building and asks the student to test a solution, emphasizing student self-completion.
Therefore, we ask a research question: \textit{``how do we capture differences in tutoring styles and steer LLMs to exhibit these styles?''} 

\subsection{Contributions}
In this paper, we explore how implicit tutor personas embedded in human tutor-student dialogues can be used to guide LLM behavior, rather than relying on explicitly prompted instructions. 
We use preference optimization to learn a steering vector and a scaling coefficient that capture and control the persona direction in the model’s representation space. Using these learned parameters, we then steer the LLM in a tutor-specific manner.
We find that our approach enables consistent recovery of tutor-specific steering across dialogue contexts, shifting the model away from population-mean behavior and toward the behavior exhibited by individual tutors. We further show the strength of this effect can be controlled at inference time through a global scaling factor, allowing for smooth adjustment of the intervention. Finally, analysis of the learned scaling coefficients reveals an interpretable structure over tutors that corresponds to variation in tutoring behavior, supporting activation steering as an effective way for capturing and controlling tutor-specific variation.

\section{Related Work}

\subsection{Tutoring Strategies}
Effective tutoring relies on a set of practical strategies rather than a single teaching approach, with tutors adjusting what they do in response to the learner’s needs. One core strategy is scaffolding, in which tutors provide support that helps learners solve problems they could not complete on their own, and then gradually reduce that support as learners become more capable \cite{wood1976role}. Tutors also vary how directly they teach, moving between clear explanation or demonstration and guided practice that helps learners take on more responsibility over time \cite{rosenshine2012principles}. In tutoring conversations, questioning strategies are used to prompt learners to explain their thinking while keeping the interaction focused on the learning goal \cite{graesser1995collaborative}. During early stages of learning, tutors often use worked examples to show how problems can be solved and to make new skills easier to grasp \cite{sweller1985use}. As learners practice, tutors provide formative, error-focused feedback to identify misunderstandings and support improvement \cite{shute2008focus}. Finally, effective tutoring includes affective strategies such as encouragement, building rapport, and managing frustration in order to maintain engagement and persistence \cite{d2012dynamics}.

\subsection{Embedding Personas in LLMs}
Research on embedding personas into LLMs generally follows two approaches: text-based conditioning and data-driven learning. In the text-based approach, personas are explicitly specified using structured descriptors such as the Five-Factor Model (Big Five)~\cite{mccrae1992introduction}, with prior work demonstrating that prompting an LLM with predefined Big Five trait profiles can induce consistent personality-aligned behaviors, which are validated by administering standardized personality questionnaires to the model \cite{serapio2023personality}. In the context of intelligent tutoring systems, related work adopts explicit persona specifications to construct personality-aware student simulations that combine cognitive attributes and personality traits, enabling systematic analysis of how different learner personas elicit and evaluate adaptive tutoring strategies \cite{liu2024personality}.

In contrast, data-driven learning induces personas from examples rather than explicit textual descriptions, and is commonly implemented either through SFT or by manipulating representation-level features~\cite{chen2025persona,mazare2018training}. In this paper, we focus on activation steering using steering vectors, which estimate a direction from contrastive behaviors and inject that direction into a model’s hidden states to bias generation toward a target persona. A steering vector can be computed from differences in internal activations produced by paired inputs or outputs that reflect desired versus undesired traits, and then applied at selected layers and token positions during the forward pass to produce consistent behavioral shifts~\cite{liu2023context,rimsky2024steering,turner2023steering,wang2023trojan}. Our work follows BiPO, which itself builds on Direct Preference Optimization (DPO) by combining preference-based learning to train a steering vector with activation steering to achieve finer-grained control over model behavior~\cite{cao2024personalized,rafailov2023direct}. Like DPO, BiPO uses relative preferences between model outputs rather than absolute likelihood targets, yielding a contrastive learning signal that enables bidirectional steering between preferred and non-preferred behaviors.

\section{Methodology}

In this section, we detail our approach to learning a steering vector from tutor-student dialogues. We introduce the notation used throughout the paper:
\[
\begin{aligned}
\mathcal{D} &\triangleq \bigcup_{i=1}^{I} \{(q_j,d_j^i)\}_{j=1}^{J_i},\\
d_j^i &\triangleq (s_{j,0},t_{j,1}^i,s_{j,1},\ldots,t_{j,K_j}^i,s_{j,K_j}).
\end{aligned}
\]
\(\mathcal{D}\) denotes a human tutor-student dialogue dataset, where \(I\) is the total number of tutors and \(i \in \{1,\ldots,I\}\) indexes a specific tutor. For each tutor \(i\), \(J_i\) denotes the number of dialogues conducted by tutor \(i\), and \(j \in \{1,\ldots,J_i\}\) indexes an individual dialogue. Each dialogue \(d^i_j\) is associated with a math question \(q_j\). 
%
Each dialogue \(d^i_j\) consists of an alternating sequence of student turns \(s_{j,k}\) and tutor turns \(t^i_{j,k}\). \(k \in \{1,\ldots,K_j\}\) indexes (student, tutor) turn pairs in the dialogue, where \(K_j\) is the total number of such turn pairs. The initial student turn \(s_{j,0}\) is included when the student initiates the dialogue. We refer to the textual content of each turn as an \emph{utterance}.

A steering vector encodes a direction in an LLM's activation space, and when added to the activations induces changes in the LLM's behavior. Learning a tutor-specific vector requires making explicit what the model is being steered away from and toward. In our setting, we define this direction relative to an \emph{average} tutor: for a given dialogue context, the steering vector should shift the model from a population-mean tutor utterance toward the utterance produced by a particular tutor. 
We first SFT the LLM to condition on the dialogue context and generate an appropriate tutor response. The adaptation ensures that the model can perform the tutoring-response generation task before we estimate tutor-specific directions in activation space. We then use the fine-tuned model to generate a population-mean tutor utterance for each dialogue context and pair this generated utterance with the corresponding ground-truth tutor utterance. Finally, we learn a steering vector from these paired examples, so that the resulting vector captures tutor-specific stylistic variation beyond the population-average tutoring behavior.
See Appendix Algorithm~\ref{alg:our_impl} for overall procedure to learn the steering vector.
%

\subsection{Constructing Pairs}
We use SFT to fine-tune an LLM to serve as a proxy for population-mean behavior, reducing response diversity and encouraging convergence toward a generic average-tutor behavior~\cite{li2024preserving}; the resulting model is used to generate population-mean utterances.
Let $\pi_{\theta}$ denote an LLM parameterized by $\theta$; we train $\pi_{\theta}$ via SFT by minimizing the negative log-likelihood of tutor utterances in $\mathcal{D}$ conditioned on the associated question $q_j$ and the preceding dialogue context $(s_{j,\le k},\, t^i_{j,<k})$:
\[
\frac{1}{I}\sum_{i=1}^{I}
\frac{1}{J_i}\sum_{j=1}^{J_i}
\frac{1}{K_j}\sum_{k=1}^{K_j}
-\!\log\!\pi_{\theta}(
t^i_{j,k}\!\mid\!s_{j,\le k},t^i_{j,<k},q_j
).
\]
For the preceding dialogue context, we construct a pair in which the ground-truth tutor utterance is the preferred response and a response generated by the SFT model is the dispreferred response. The SFT-generated response serves as a proxy for population-mean tutor behavior, so each pair explicitly defines the direction from the average tutor response \(\bar t_{j,k}\) toward the tutor-specific response \(t_{j,k}^i\). See Appendix Tables~\ref{tab:appendix_drastic_diff_examples_1} and~\ref{tab:appendix_drastic_diff_examples_2} for examples of such contrastive pairs.

\subsection{Learning a Steering Vector}
We now detail what a steering vector is and how we learn it using the pairs.
We introduce a steering vector $v$, shared across tutors, which we use to modify the model’s activation during inference by adding it to the latent activations $A_L(\cdot)$ at layer $L$. We assign each tutor a positive scaling coefficient \(\delta_i\) that scales a shared steering direction \(v\). We constrain these coefficients to have unit mean across tutors to eliminate a scale ambiguity, since the intervention depends only on the product \(\delta_i v\) and could otherwise be left unchanged by scaling \(\delta_i\) up while scaling \(v\) down (or vice versa) without affecting the loss. To parameterize \(\delta_i\) while enforcing positivity, we introduce a tutor-specific learnable scalar \(u_i\) and define $\delta_i = \exp(u_i)\big/\big(\frac{1}{I}\sum_{m=1}^{I}\exp(u_m)\big)$. We jointly optimize each \(u_i\) and the shared steering direction \(v\) in training. We initialize $u_i = 0$ for all tutors ($\delta_i = 1$), and initialize $v$ from a standard Gaussian scaled by $0.5$ to ensure a sufficiently strong initial steering magnitude. We note that while we learn the steering vector through preference optimization inspired by BiPO~\cite{cao2024personalized}, we differ in that, rather than using bidirectional steering, we learn a scaling coefficient that scales a shared steering direction:
\[
\min_{v,u}\;
\mathbb{E}_{i,j,k}\!\left[
-\log \sigma\!\left(
\beta \log \frac{
\pi_{\theta}(t^i_{j,k} \mid A_L(\cdot) + \delta_i v)
}{
\pi_{\theta}(t^i_{j,k} \mid A_L(\cdot))
}
\right.\right.
\]
\[
\left.\left.
- \beta \log \frac{
\pi_{\theta}(\bar t_{j,k} \mid A_L(\cdot) + \delta_i v)
}{
\pi_{\theta}(\bar t_{j,k} \mid A_L(\cdot))
}
\right)\right].
\]
For tutor $i$, dialogue $j$, turn $k$, our objective includes two relative preference terms. The first log term encourages the steered LLM to have higher likelihood in generating actual tutor utterance $t^i_{j,k}$ compared to the unsteered SFT model. The right term does the exact opposite, encouraging the steered model to have lower likelihood in generating the population-average tutor utterance. 
All likelihoods in these terms are sequence log-likelihoods computed by summing
next-token log-probabilities over the response tokens only, with prompt tokens
masked out.
The logistic loss encourages learning a shared steering direction $v$ and tutor-specific coefficients $\delta_i$ such that applying $\delta_i v$ consistently favors tutor-specific outputs across tutors, dialogues, and turns. The hyperparameter $\beta$ encourages similarity to the base model's behavior without the steering vector applied. We optimize this objective by iterating for $T$ steps until the loss converges.

At test time, we introduce a global scaling factor $\alpha$ and apply activation steering as $A_L(\cdot)\leftarrow A_L(\cdot)+\alpha\, \delta_i v$. 
We observe that $\alpha=1$ leads to longer utterances, because of the objective
uses unnormalized sequence log-likelihoods, so the log-ratio accumulates over all response tokens. Consequently, longer tutor utterances can contribute larger preference margins than shorter ones, making the learned intervention at $\alpha=1$ overly strong on continuing generation rather than stopping.
While the training objective identifies a shared steering direction $v$ and relative tutor-specific strengths $\delta_i$ with respect to the unsteered SFT model, a separate scaling factor is necessary to control the overall strength of the intervention during inference.

\section{Experimental Settings}
\subsection{Dataset}
We use the Question-Anchored-Tutoring-Dialogues-2k dataset~\cite{eedi-dialogues}, containing dialogues from web-based conversations between expert tutors and real students in an online math learning platform. The conversations are initiated by tutors or students requesting assistance while working through a lesson. 
We split the dataset into training, validation, and test sets at the dialogue level using an 80/10/10 split for each of the 21 unique tutors. On average, each tutor has 74.19 ($\sigma = 75.47$) dialogues in the training set, 8.90 ($\sigma = 9.45$) dialogues in the validation set, and 10.10 ($\sigma = 9.42$) dialogues in the test set. The average number of turns per dialogue is 11.75 ($\sigma = 1.10$) in training, 12.06 ($\sigma = 2.43$) in validation, and 12.25 ($\sigma = 1.73$) in test. 

\subsection{Evaluation Setup and Metrics}
\label{sec:setup}

We evaluate how closely the steered tutor utterances match the corresponding ground-truth on a per-tutor basis, along two main axes: \emph{lexical similarity} and \emph{semantic similarity}. For each tutor $i$, we compare steered outputs $t_{j}^{\delta_i}$ to ground-truth utterances $t_j$ across turns $j$. For lexical similarity, we measure surface-form overlap using ROUGE-L (\textbf{R})~\cite{lin2004rouge} and BLEU(\textbf{B})~\cite{papineni2002bleu}. For semantic similarity, we assess semantic alignment using cosine similarity in embedding space (\textbf{C}) and a pairwise preference-based win rate (\textbf{W}). We define
$\mathrm{C}(\delta_i) = \mathbb{E}_{j}\!\left[\cos\!\big(\phi(t_{j}^{\delta_i}), \phi(t_j)\big)\right]$,
where $\phi(\cdot)$ denotes a sentence embedding function from a pretrained \texttt{SentenceBERT} encoder~\cite{reimers2019sentence}.
For Win Rate, we use LLM-as-a-judge with \texttt{Prometheus-Eval}, given a dialogue and a ground-truth turn and chooses which of the steered or unsteered utterances more closely aligns with the tutoring style and language used~\cite{kim2024prometheus}. We report the win rate over unsteered SFT utterances, where $0.5$ indicates the two perform equally. We do not report ties because they are rare. All metrics are computed per tutor and then averaged across tutors.

\subsection{Model Parameters}
We use \texttt{Llama-3.1-8B-Instruct} as the base model for SFT, which we adapt using LoRA~\cite{hu2021lora} with a learning rate of $5\times10^{-5}$ over $3$ epochs, setting the LoRA rank to $r=32$ and the scaling factor to $\alpha=64$. Our SFT setup follows \citet{scarlatos2026simulated}, adopting their prompts and training procedure.
For learning $v$ and $u_i$, we use the validation set, applying each $u_i$ to the dialogues corresponding to tutor $i$ and selecting hyperparameters by tracking validation loss; we set $\beta=1.0$ and learning rate $=0.01$, and apply activation steering at the final transformer layer (layer 32).
Text generation uses a temperature of $1.0$ and top-$p$ sampling with $p=0.95$, yielding a final validation loss of $0.543$ at $T=17$. The learned $u_i$ have a mean of $0.053$ and a standard deviation of $0.106$. The activation steering mechanism follows the persona vector approach~\cite{chen2025persona}, where the learned steering vector is added to the layer activations at all token positions. All training and evaluation are conducted on a single NVIDIA A40 GPU with 48\,GB of VRAM.

\section{Experimental Results}
\label{sec:result}

\subsection{Quantitative Results}

We quantitatively evaluate activation steering for tutoring along two dimensions: (1) how well it recovers tutor-specific behavior at different stages of a dialogue, and (2) how this behavior varies with the strength of the steering intervention. Specifically, we first compare steered tutor utterances $\hat{t}_j$ with population-mean utterances $\bar{t}_j$, using ground-truth tutor utterances $t_j$ as reference, in different stages of the dialogue (Table~\ref{tab:quant_across}). We then analyze the effect of varying the global scaling factor $\alpha$ on both stylistic alignment and lexical similarity (Table~\ref{tab:quant_alpha}).

We conduct a stage-wise analysis because different parts of a dialogue serve different purposes. Early and late turns are largely greetings, encouragement, and closing remarks, which tend to follow common conventions and thus provide limited signal about a tutor’s instructional style. In contrast, the middle portion of the dialogue contains the core problem-solving interaction, where tutors explain concepts, provide hints, and respond to student errors. These turns are where tutor-specific strategies and language use are most evident. For this reason, we define the \emph{mid stage} as the 10--90\% interval of turns within each dialogue and focus our analysis on this region. 


\begin{table}[t]
\caption{Comparison of population-mean utterances $\bar{t}_j$ and steered tutor utterances $\hat{t}_j$ against the ground-truth tutor utterances $t_j$ across dialogue stages. Metrics are computed per tutor and then averaged across tutors.}
\centering
\small
\setlength{\tabcolsep}{4pt}
\begin{tabular}{ccccccc}
\toprule
\textbf{Stage} & \textbf{Count} & \textbf{Method} &
\textbf{R$^\uparrow$} & \textbf{B$^\uparrow$} &
\textbf{C$^\uparrow$} & \textbf{W$^\uparrow$} \\
\midrule
\multirow{2}{*}{early}
&  \multirow{2}{*}{326}  & $\bar{t}_j$
& 0.285 & 0.070 & 0.392 & -- \\
\cmidrule(lr){3-7}
&  & $\hat{t}_j$
& 0.206 & 0.041 & 0.321 & 0.571 \\
\midrule
\multirow{2}{*}{mid}
&  \multirow{2}{*}{1971}  & $\bar{t}_j$
& 0.165 & 0.019 & 0.385 & -- \\
\cmidrule(lr){3-7}
&  & $\hat{t}_j$
& 0.157 & 0.018 & 0.426 & 0.587 \\
\midrule
\multirow{2}{*}{late}
&  \multirow{2}{*}{326} & $\bar{t}_j$
& 0.157 & 0.025 & 0.330 & -- \\
\cmidrule(lr){3-7}
&  & $\hat{t}_j$
& 0.121 & 0.017 & 0.349 & 0.564 \\
\bottomrule
\end{tabular}
\label{tab:quant_across}
\end{table}

\begin{table}[t]
\caption{Comparison of varying the strength of the steering vector by applying a scaling factor $\alpha$ in activation steering for all stages: $A_L(\cdot)\leftarrow A_L(\cdot)+\alpha\,\delta_i v$.}
\label{tab:quant_alpha}
\centering
\small
\setlength{\tabcolsep}{4pt}
\begin{tabular}{ccccccc}
\toprule
\textbf{Count} & \textbf{Method} & \textbf{$\alpha$} &
\textbf{R$^\uparrow$} & \textbf{B$^\uparrow$} &
\textbf{C$^\uparrow$} & \textbf{W$^\uparrow$} \\
\midrule
\multirow{5}{*}{2623}
& $\bar{t}_j$ & 0.0 
& 0.179 & 0.026 & 0.379 & -- \\
\cmidrule(lr){2-7}
& \multirow{4}{*}{$\hat{t}_j$} & 0.3
& 0.181 & 0.027 & 0.400 & 0.536 \\
&  & 0.5
& 0.187 & 0.028 & 0.407 & 0.539 \\
&  & 0.7
& 0.176 & 0.026 & 0.404 & 0.562 \\
&  & 1.0
& 0.159 & 0.021 & 0.403 & 0.582 \\
\bottomrule
\end{tabular}
\end{table}

Table~\ref{tab:quant_across} shows the results on all metrics. Beyond early dialogue stages, steering leads to a clear improvement in semantic alignment with the ground-truth tutor: in the mid stage, cosine similarity increases from 0.385 for $\bar{t}_j$ to 0.426 for $\hat{t}_j$, and the LLM judge prefers $\hat{t}_j$ over the unsteered baseline in 58.7\% of comparisons. At the same time, lexical overlap is only slightly reduced; ROUGE-L decreases from 0.165 to 0.157 and BLEU remains almost unchanged (0.019 to 0.018). This result suggests that steering improves tutor-specific style primarily by altering phrasing and discourse structure rather than by matching exact word sequences. Only in the early stage, the cosine similarity for $\hat{t}_j$ is worse than $\bar{t}_j$. The result is likely due to greetings in the early turns being inherently generic, which means that adding the steering vector over-exaggerates the effect of tutor style in these turns. A solution may be dynamically setting $\alpha$ to lower values for turns that are intended to be more generic.

Table~\ref{tab:quant_alpha} examines the effect of steering strength aggregated across all dialogue stages. Moderate values of $\alpha$ ($0.3$ and $0.5$) slightly improve ROUGE-L and BLEU relative to the population-mean baseline, while also increasing cosine similarity. As $\alpha$ increases, Win Rate rises steadily from 0.536 to 0.582, indicating stronger expression of tutor-specific behavior. However, higher values of $\alpha$ also lead to lower ROUGE-L and BLEU scores, reflecting a trade-off in which stronger steering enhances stylistic alignment at the cost of lexical similarity.

Based on these results, we select $\alpha = 0.5$ for our qualitative analysis, since it provides a good balance between improving semantic and stylistic alignment with the target tutor while largely preserving lexical similarity.

\subsection{Qualitative Results}

\begin{figure*}[t]
    \centering
    \includegraphics[width=1\linewidth]{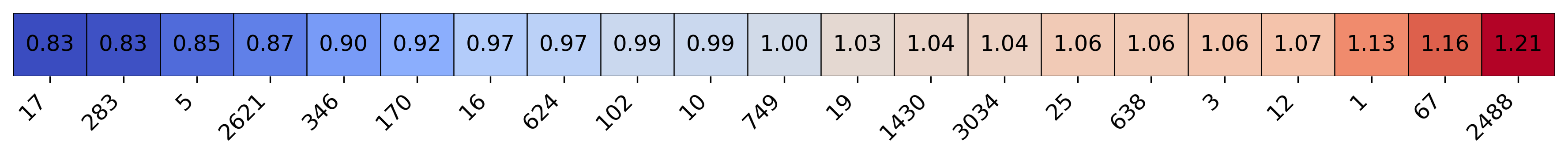}
    \vspace{-.5cm}
    \caption{Heatmap of $\delta_i$ across $21$ tutors with different tutoring styles.}
    \label{fig:heatmap}
\end{figure*}

\subsubsection{Interpreting the steering direction}

We visualize the scaling coefficient $\delta_i$ in Figure~\ref{fig:heatmap}. To examine how $\delta_i$ correlates with tutoring styles or language use, we put tutors into three groups that each address a common math question in the dataset\footnote{Question IDs: 104637, 91697, and 77628, respectively.}: from the left end to the middle (Tutors 283, 346, and 102), from the middle to the right (Tutors 283 and 1430), and from the middle to the right end (Tutors 19 and 2488). Table~\ref{tab:problems_and_gt} provides an overview of the math questions and ground-truth utterances, where tutors with lower $\delta_i$ are placed on the left and those with higher $\delta_i$ are placed on the right.

Table~\ref{tab:problems_and_gt} shows three example math questions (ratios, linear sequences, and order of operations) together with short tutor-student dialogue snippets. The table is organized to make the learned direction $\delta_i$ interpretable: tutors are arranged from left to right according to their position in the heatmap ordering (e.g., $283 \rightarrow 346 \rightarrow 102 \rightarrow 1430 \rightarrow 19 \rightarrow 2488$). For each question, the ``Lower $\delta_i$'' column contains tutors closer to the left side of this ordering, while the ``Higher $\delta_i$'' column contains tutors closer to the right side. The blue-highlighted text marks the target tutor utterance to be predicted. Therefore, snippets show how the same dialogue context elicits different conversational moves and tutor utterances, as we move along the $\delta_i$ direction.

\begin{table*}[t!]
\centering
\caption{Examples of tutor-student dialogue across different math topics. \textcolor{blue!60!black}{Blue} denotes our target utterance.}
\label{tab:problems_and_gt}
\small
\setlength{\tabcolsep}{6pt}
\renewcommand{\arraystretch}{1.15}
\scalebox{.9}{
\begin{tabular}{@{} l l l @{}}
\toprule
\textbf{Question} & \textbf{Lower $\delta_i$} & \textbf{Higher $\delta_i$} \\
\midrule

\parbox[t]{0.2\textwidth}{
\textbf{Ratios}\\
\textit{Lila and Elijah share some sweets in the ratio $3:2$.\\
What is this ratio in the form $1:n$?\\[2pt]}
A.\ $1:1.5$ \\ B.\ $1:\frac{2}{3}$\\
C.\ $1:0$ \\ D.\ $1:6$
}
&
\parbox[t]{0.42\textwidth}{
\textbf{Tutor 283}\\
\texttt{T}: :-) So, you need to find an equivalent ratio to $3:2$. That is, instead $1:n$ \\
\texttt{S}: im not sure \\
\texttt{T}: So, what would you divide 3 by, to get the answer 1? \\
\texttt{S}: 3 divide by 3\\
\texttt{T}: \textcolor{blue!60!black}{Yes, great! So, the 3 in the ratio has been divided by 3, so you now need to divide the 2 by 3 as qwell. *well}
}
&
\parbox[t]{0.42\textwidth}{
\textbf{Tutor 102}\\
\texttt{T}: Ok let's take a look. What do we need to divide by to get the start of the ratio down from 3 to 1? \\
\texttt{S}: i dont know \\
\texttt{T}: $3 \div \_\_\_ = 1$ \\
\texttt{S}: 1 \\
\texttt{T}: \textcolor{blue!60!black}{Almost, try again?}
}
\\
\midrule
\parbox[t]{0.2\textwidth}{
\textbf{Linear \\sequence}\\
\textit{How many dots would there be in Pattern \( n \) ? [Image: 6 dots, 10 dots, 14, and 18 dots]
\\[2pt]}
A.\ $4 n+2$ \\ B.\ $4 n-2$\\
C.\ $2 n+4$ \\ D.\ $n+2$
}
&
\parbox[t]{0.42\textwidth}{
\textbf{Tutor 283} \\
\texttt{T}: Yes of course, let's have a look So, how many dots are being added each time? \\
\texttt{S}: 4? \\
\texttt{T}: Great! So that tells us that the first part of our formula will be 4n \\
\texttt{S}: so does the 4n refer to how many dots are being added each time? \\
\texttt{T}: \textcolor{blue!60!black}{Yes, that's how we work out that bit Then, we need to see if that works, or if we need to add/subtract anything When n=1, 4n = ?} \\
}
&
\parbox[t]{0.42\textwidth}{
\textbf{Tutor 1430} \\
\texttt{T}: Great, now pattern 2? \\
\texttt{S}: 10 \\
\texttt{T}: Great, now pattern 3? \\
\texttt{S}: 14 \\
\texttt{T}: So what is the pattern going up in? \\
\texttt{S}: 4 \\
\texttt{T}: \textcolor{blue!60!black}{6, 10, 14,} \\
}
\\
\midrule
\parbox[t]{0.2\textwidth}{
\textbf{Order \\ of operations}\\
\textit{What is the value of\\
$\frac{1+3 \times 4}{2}$?\\[2pt]}
A.\ $12.5$ \\ B.\ $8$\\
C.\ $7$ \\ D.\ $6.5$
}
&
\parbox[t]{0.42\textwidth}{
\textbf{Tutor 19} \\
\texttt{T}: Yep no worries With a fraction like this we would need to calculate the value of the numerator first What is 1 + 3 x 4 \\
\texttt{S}: well 1 add 3 is 4 and 4 times 4 is 16 so will it be 16< \\
\texttt{T}: Not quite. Remember with the order of operations we multiply before we add so we need to do 3 x 4 first then add 1 \\
\texttt{S}: oh 3 times 4 is 12 add the 1 is 13 \\
\texttt{T}: \textcolor{blue!60!black}{\raisebox{-0.1em}{\includegraphics[height=1em]{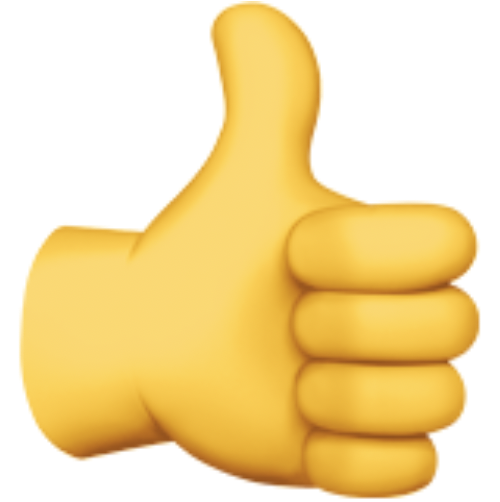}} So now we have 13/2 Remember the fraction line means divide}
}
&
\parbox[t]{0.42\textwidth}{
\textbf{Tutor 2488} \\
\texttt{S}: What do I do after 3x4 \\
\texttt{T}: With the fraction line what does that actually mean? \\
\texttt{S}: I’m not sure what it means \\
\texttt{T}: It means divide \\
\texttt{S}: So 12 divided by 2 \\
\texttt{T}: \textcolor{blue!60!black}{So it's a weird one because you'd normally do division before the addition but because it's all of the 1+12 divided by 2} \\
}

\\
\bottomrule
\end{tabular}
}
\end{table*}

Tutor~283 who appears at the left end shows relational and pedagogical investment through explicit encouragement and step-by-step scaffolding. In the ratios example, Tutor~283 uses affective markers and praise (``:-)'', ``Yes, great!'') and then connects the student’s correct operation on the first term to the required operation on the second term: ``the 3 in the ratio has been divided by 3, so you now need to divide the 2 by 3 as well.'' In the linear sequence example, the tutor validates the student (``Great!''), explains the structural role of $4n$, and checks understanding with a concrete substitution prompt (``When $n=1$, $4n = ?$''). These turns prioritize both student confidence and conceptual clarity.

Tutors positioned toward the middle of the $\delta_i$ direction, including Tutors~102, 1430, and~19, share a more efficiency-oriented style. Across examples, these tutors rely on short, task-focused prompts and corrective feedback, with reduced affective language and limited elaboration. Tutor~102 narrows the student’s response space through a fill-in-the-blank equation (``$3 \div \_\_\_ = 1$'') and provides brief correction (``Almost, try again?''). Tutor~1430 advances the linear sequence problem through a series of concise checkpoints (``Great, now pattern~2?'', ``Great, now pattern~3?'') and a minimal continuation of the sequence (``6, 10, 14,''). Tutor~19 similarly emphasizes efficient progression by providing succinct procedural guidance (``we multiply before we add'') and a short conceptual reminder (``Remember the fraction line means divide''), occasionally accompanied by a small positive signal with thumbs up emoji. Together, these tutors demonstrate consistent pedagogical engagement while prioritizing error diagnosis and progress over extended explanation or rapport-building.

At the right end of the $\delta_i$ direction, Tutor~2488 closely aligns with a low-investment style, focusing on just getting the task completed. The tutor gives short answers (``It means divide'') and follows up with a confusing justification (``it’s a weird one because you’d normally do division before the addition''), which does not clearly resolve the student’s uncertainty about expression structure. Compared to Tutor~19’s concise but accurate guidance, Tutor~2488’s response prioritizes moving the interaction forward, even at the expense of clarity in their explanation. Taken together, these examples indicate that the learned direction $\delta_i$ corresponds to changes in tutoring style, from rapport-rich scaffolding on the left, through diagnostic instruction in the middle, to procedural assistance that emphasizes task completion over teaching on the right.

\subsubsection{Case studies}
Table~\ref{tab:gen} shows predicted utterances from unsteered SFT and steered generations with $\alpha = 0.5$. Overall, steering makes the generations better aligned with the ground-truth tutor utterances in Table~\ref{tab:problems_and_gt}, particularly in how the LLM matches the tutoring approach associated with $\delta_i$. In many cases, the unsteered model produces reasonable tutoring moves, but it often misses the specific dialogue act or tone that distinguishes lower-$\delta_i$ tutors (rapport and scaffolding) from higher-$\delta_i$ tutors (corrective or short/concise continuations). With $\alpha = 0.5$, the model more consistently reproduces the intended style difference between the two ends, yielding utterances that are closer to the ground-truth utterances in both tutoring style and language use.

\begin{table*}[t!]
\small
\caption{Predicted utterances for the target utterances shown in Table~\ref{tab:problems_and_gt}.}
\label{tab:gen}
\setlength{\tabcolsep}{6pt}
\renewcommand{\arraystretch}{1.15}
\scalebox{.9}{
\begin{tabular}{@{} l p{0.51\textwidth} p{0.51\textwidth} @{}}
\toprule
$\boldsymbol{\alpha}$ & \textbf{Lower $\delta_i$} & \textbf{Higher $\delta_i$} \\
\midrule

$0.0$ & Amazing!! And 2 divide by 3? & So we need to divide both sides by 3 3÷3=1 What would the ratio become? \\
$0.5$ & And what would happen to 2? \raisebox{-0.2em}{\includegraphics[height=1em]{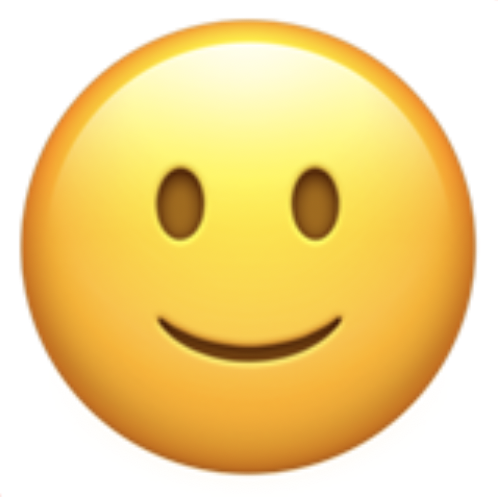}} 3÷3 = 1 So, we would have to divide both parts by 3 2÷3 =? & Nearly, not 1... What do we need to divide by to get from 3 to 1? \\
\midrule

$0.0$ & Exactly, so if it's pattern 1 it will have 4n dots in the pattern & Great, so to get pattern n, we have 4(n) How many is it missing? \\
$0.5$ & 4n refers to the number of times 4 is being added So, 4n means the number of terms we're adding 4 to And then, we have to account for the fact that in the first pattern (n=1) there are 6 dots, not 4 So, we start with 4n and then add 2 & So in pattern 1 how many were there and what is this times by?? \\
\midrule

$0.0$ & So when we have 1 + 3 x 4, this is different from 3 x (1 + 4) Are you ok with why that's different? & You have to carry out both operations \\
$0.5$ & brilliant Now we have 13 / 2 & Yes \raisebox{-0.1em}{\includegraphics[height=1em]{thumbs_up.png}} \\

\bottomrule
\end{tabular}
}
\end{table*}

In the ratios example, the ground-truth utterance of Tutor~283 first affirms the student and then explains that whatever is done to the 3 must also be done to the 2. The unsteered prediction (``Amazing!! And 2 divide by 3?'') has a friendly tone, but moves directly to $2/3$ without introducing additional steps. The steered prediction, however, provides a conceptual step while maintaining a friendly tone, thereby exhibiting a tutoring style similar to the ground truth. The ground-truth utterance of Tutor~102 is a brief correction (``Almost, try again?'') following an incorrect student answer. The unsteered prediction instead shifts into an elaborated instructional explanation (``So we need to divide both sides by 3 ... What would the ratio become?''), which does not match Tutor~102’s concise corrective move. The steered prediction  ``Nearly, not 1... What do we need to divide by to get from 3 to 1?'' is much closer in style: it is explicitly corrective (``Nearly, not 1...'') and prompts the student to retry the same subproblem, matching the ground-truth tutor strategy even if the exact wording differs.

In the linear sequence example, there is a similar pattern. The ground-truth utterance of Tutor~283 answers the student’s clarification question about what $4n$ represents and then immediately transitions into a structured check (``When $n=1$, 4n = ?''). The unsteered prediction (``Exactly, so if it's pattern 1 it will have 4n dots ...'') moves the problem forward but does not directly address the student’s question, which differs from the tutor’s explanatory move. The steered prediction shows an explicitly explanatory turn that attempts to clarify what $4n$ means and then addresses it by referencing the mismatch at $n=1$, introducing an offset (``we start with 4n and then add 2''). While the steered text is more verbose and uses some slightly awkward phrasing, it is closer in tutoring strategy because it responds to the student’s conceptual query and uses a concrete consistency check to motivate the adjustment, which is the same pedagogical move that Tutor~283 uses. The ground-truth utterance of Tutor~1430 is extremely concise (``6, 10, 14,''), functioning more like a checkpoint or continuation than an explanation. The unsteered prediction introduces additional scaffolding (``How many is it missing?''), and the steered prediction becomes even more diagnostic (``So in pattern 1 how many were there and what is this times by??''). Neither matches the concise tutoring style in the ground truth, but the steered utterance aligns better in that it remains concise and task-driven.

In the order of operations example, steering yields the clearest improvement. The ground-truth utterance of Tutor~19 ends with a short affirmation with a procedural reminder, ``(:thumbs up) So now we have 13/2 Remember the fraction line means divide.'' The unsteered prediction instead introduces a conceptual aid, contrasting two different expressions (``this is different from 3 x (1 + 4)''), which is pedagogically plausible but does not match the ground-truth target’s concise confirmation-and-next-step structure. The steered prediction shows ``brilliant Now we have 13 / 2,'' which is substantially closer in both tone and function, echoing the short praise and the immediate transition to the next step. The ground-truth utterance of Tutor~2488 is a long, somewhat muddled explanation about why the fraction is ``weird,'' a term rarely used to explain order of operations. Therefore, both predictions struggle, and the steered one incorrectly assesses the student response as correct. The unsteered prediction is overly generic while the steered shows an acknowledgment, saying yes. 

Across all examples, steering consistently improves alignment between model-generated utterances and ground-truth ones, capturing the specific tutoring strategy, tone, and style along the $\delta_i$ spectrum that unsteered SFT often misses. With $\alpha = 0.5$, steered outputs more reliably reflect the intended tutoring stance and underlying tutoring function, such as rapport-building and scaffolding, affirming before explaining, or concise, corrective continuations, suggesting steering effectively controls pedagogical intent beyond surface style.

\newpage
\section{Conclusions and Future Work}

In this paper, we explored how implicit signals in human tutor-student dialogues can be used to guide LLMs towards different tutoring styles. We proposed an activation steering method that learns a shared steering direction and tutor-specific coefficients from real tutor-student dialogues, relative to a population-mean reference, enabling personalized, controlled tutor-specific behavior. We show that the learned steering direction and strengths lead to improved replication of individual tutor behavior. We also qualitatively show that it encodes changing tutoring styles from scaffolded, rapport-building through efficient, task-focused guidance to minimal, task-completion-oriented instruction.

There are several directions for future work. First, although our objective captures tutor-specific behavior via preferences, it does not explicitly encode pedagogical principles. Integrating pedagogical rubrics, learning-theoretic goals, or student outcome-based rewards could better align models with effective tutoring practices. Second, beyond steering tutor behavior, future work could learn complementary steering vectors for student models, enabling simulated tutor–student interactions driven by latent student personas. This would allow analysis of how interaction dynamics and tutor–student compatibility influence learning outcomes. Finally, extending the method to learn multiple steering directions, for example through low-rank subspaces from dimensionality reduction, could enable more fine-grained and disentangled control over distinct pedagogical attributes.

\section*{Acknowledgments} This work is partially supported by Renaissance Philanthropy under the learning engineering virtual institute (LEVI) initiative and the NSF under grants 2153481 and 2237676. 

\section*{Limitations}

There are several practical limitations to our work. First, we evaluate our method on a single math dialogue dataset, so it is unclear how well the results generalize to other datasets, domains, or tutoring settings. Second, our baselines are limited to supervised fine-tuning methods; future work should compare against prompting-based approaches as well. Third, we do not include human evaluation, so our findings rely on automatic metrics and model-based judgments.

\section*{Ethical Considerations}

Our experiments use the Eedi dataset~\cite{eedi-dialogues}, in which dialogues are associated with pseudonymous identifiers rather than real names. We therefore do not identify any individual tutors or students in this work. Beyond this, we do not foresee additional ethical concerns specific to our experimental setup.

\bibliography{custom}

@article{cao2024personalized,
  title={Personalized steering of large language models: Versatile steering vectors through bi-directional preference optimization},
  author={Cao, Yuanpu and Zhang, Tianrong and Cao, Bochuan and Yin, Ziyi and Lin, Lu and Ma, Fenglong and Chen, Jinghui},
  journal={Advances in Neural Information Processing Systems},
  volume={37},
  pages={49519--49551},
  year={2024}
}

@misc{livehint,
	Howpublished = {Online: \url{https://support.carnegielearning.com/help-center/math/livehint/article/livehint-overview/}},
	Title = {LiveHint Overview},
	year={2024},
	author={{Carnegie Learning}}
}

@misc{khanmigo,
	Howpublished = {Online: \url{https://www.khanmigo.ai/}},
	Title = {Supercharge your teaching experience with Khanmigo},
	year={2023},
	author={{Khan Academy}}
}

@misc{chatgpt,
  title        = {ChatGPT},
  author       = {{OpenAI}},
  year         = {2023},
  howpublished = {\url{https://chat.openai.com}},
  note         = {Accessed: 2026-01-21}
}

@misc{gemini,
  title        = {Gemini},
  author       = {{Google}},
  year         = {2023},
  howpublished = {\url{https://deepmind.google/technologies/gemini/}},
  note         = {Accessed: 2026-01-21}
}

@inproceedings{scarlatos2025training,
  title={Training llm-based tutors to improve student learning outcomes in dialogues},
  author={Scarlatos, Alexander and Liu, Naiming and Lee, Jaewook and Baraniuk, Richard and Lan, Andrew},
  booktitle={International Conference on Artificial Intelligence in Education},
  pages={251--266},
  year={2025},
  organization={Springer}
}

@article{mccrae1992introduction,
  title={An introduction to the five-factor model and its applications},
  author={McCrae, Robert R and John, Oliver P},
  journal={Journal of personality},
  volume={60},
  number={2},
  pages={175--215},
  year={1992},
  publisher={Wiley Online Library}
}

@inproceedings{
li2024preserving,
title={Preserving Diversity in Supervised Fine-Tuning of Large Language Models},
author={Ziniu Li and Congliang Chen and Tian Xu and Zeyu Qin and Jiancong Xiao and Zhi-Quan Luo and Ruoyu Sun},
booktitle={The Thirteenth International Conference on Learning Representations},
year={2025},
}

@article{rafailov2023direct,
  title={Direct preference optimization: Your language model is secretly a reward model},
  author={Rafailov, Rafael and Sharma, Archit and Mitchell, Eric and Manning, Christopher D and Ermon, Stefano and Finn, Chelsea},
  journal={Advances in neural information processing systems},
  volume={36},
  pages={53728--53741},
  year={2023}
}

@inproceedings{eedi-dialogues,
    title = "{PII}vot: A Lightweight {NLP} Anonymization Framework for Question-Anchored Tutoring Dialogues",
    author = "Zent, Matthew  and
      Smith, Digory  and
      Woodhead, Simon",
    booktitle = "Proceedings of the 2025 Conference on Empirical Methods in Natural Language Processing",
    month = nov,
    year = "2025",
    pages = "27467--27476",
    ISBN = "979-8-89176-332-6"
}

@inproceedings{
hu2021lora,
title={Lo{RA}: Low-Rank Adaptation of Large Language Models},
author={Edward J Hu and yelong shen and Phillip Wallis and Zeyuan Allen-Zhu and Yuanzhi Li and Shean Wang and Lu Wang and Weizhu Chen},
booktitle={International Conference on Learning Representations},
year={2022}
}

@article{scarlatos2026simulated,
  title={Simulated Students in Tutoring Dialogues: Substance or Illusion?},
  author={Scarlatos, Alexander and Lee, Jaewook and Woodhead, Simon and Lan, Andrew},
  journal={arXiv preprint arXiv:2601.04025},
  year={2026}
}

@inproceedings{puech2025towards,
  title={Towards the pedagogical steering of large language models for tutoring: A case study with modeling productive failure},
  author={Puech, Romain and Macina, Jakub and Chatain, Julia and Sachan, Mrinmaya and Kapur, Manu},
  booktitle={Findings of the Association for Computational Linguistics: ACL 2025},
  pages={26291--26311},
  year={2025}
}

@inproceedings{sonkar2024pedagogical,
  title={Pedagogical alignment of large language models},
  author={Sonkar, Shashank and Ni, Kangqi and Chaudhary, Sapana and Baraniuk, Richard},
  booktitle={Findings of the Association for Computational Linguistics: EMNLP 2024},
  pages={13641--13650},
  year={2024}
}

@article{zhang2024spl,
  title={SPL: a socratic playground for learning powered by large language model},
  author={Zhang, Liang and Lin, Jionghao and Kuang, Ziyi and Xu, Sheng and Hu, Xiangen},
  journal={arXiv preprint arXiv:2406.13919},
  year={2024}
}

@inproceedings{dinucu2025problem,
    title = "From Problem-Solving to Teaching Problem-Solving: Aligning {LLM}s with Pedagogy using Reinforcement Learning",
    author = "Dinucu-Jianu, David  and
      Macina, Jakub  and
      Daheim, Nico  and
      Hakimi, Ido  and
      Gurevych, Iryna  and
      Sachan, Mrinmaya",
    booktitle = "Proceedings of the 2025 Conference on Empirical Methods in Natural Language Processing",
    month = nov,
    year = "2025",
    pages = "272--292",
    ISBN = "979-8-89176-332-6"
}

@inproceedings{pal2024autotutor,
  title={Autotutor meets large language models: A language model tutor with rich pedagogy and guardrails},
  author={Pal Chowdhury, Sankalan and Zouhar, Vil{\'e}m and Sachan, Mrinmaya},
  booktitle={Proceedings of the Eleventh ACM Conference on Learning@ Scale},
  pages={5--15},
  year={2024}
}

@inproceedings{scarlatos2024improving,
  title={Improving the validity of automatically generated feedback via reinforcement learning},
  author={Scarlatos, Alexander and Smith, Digory and Woodhead, Simon and Lan, Andrew},
  booktitle={International Conference on Artificial Intelligence in Education},
  pages={280--294},
  year={2024},
  organization={Springer}
}

@article{turner2023steering,
  title={Steering language models with activation engineering},
  author={Turner, Alexander Matt and Thiergart, Lisa and Leech, Gavin and Udell, David and Vazquez, Juan J and Mini, Ulisse and MacDiarmid, Monte},
  journal={arXiv preprint arXiv:2308.10248},
  year={2023}
}

@inproceedings{wang2023trojan,
author = {Wang, Haoran and Shu, Kai},
title = {Trojan Activation Attack: Red-Teaming Large Language Models using Steering Vectors for Safety-Alignment},
year = {2024},
isbn = {9798400704369},
publisher = {Association for Computing Machinery},
booktitle = {Proceedings of the 33rd ACM International Conference on Information and Knowledge Management},
pages = {2347–2357},
numpages = {11},
location = {Boise, ID, USA},
series = {CIKM '24}
}

@inproceedings{liu2023context,
author = {Liu, Sheng and Ye, Haotian and Xing, Lei and Zou, James},
title = {In-context vectors: making in context learning more effective and controllable through latent space steering},
year = {2024},
publisher = {JMLR.org},
booktitle = {Proceedings of the 41st International Conference on Machine Learning},
articleno = {1309},
numpages = {21},
location = {Vienna, Austria},
series = {ICML'24}
}

@inproceedings{rimsky2024steering,
  title={Steering llama 2 via contrastive activation addition},
  author={Rimsky, Nina and Gabrieli, Nick and Schulz, Julian and Tong, Meg and Hubinger, Evan and Turner, Alexander},
  booktitle={Proceedings of the 62nd Annual Meeting of the Association for Computational Linguistics (Volume 1: Long Papers)},
  pages={15504--15522},
  year={2024}
}

@inproceedings{papineni2002bleu,
  title={Bleu: a method for automatic evaluation of machine translation},
  author={Papineni, Kishore and Roukos, Salim and Ward, Todd and Zhu, Wei-Jing},
  booktitle={Proceedings of the 40th annual meeting of the Association for Computational Linguistics},
  pages={311--318},
  year={2002}
}

@inproceedings{lin2004rouge,
  title={Rouge: A package for automatic evaluation of summaries},
  author={Lin, Chin-Yew},
  booktitle={Text summarization branches out},
  pages={74--81},
  year={2004}
}

@inproceedings{reimers2019sentence,
    title = "Sentence-{BERT}: Sentence Embeddings using {S}iamese {BERT}-Networks",
    author = "Reimers, Nils  and
      Gurevych, Iryna",
    booktitle = "Proceedings of the 2019 Conference on Empirical Methods in Natural Language Processing and the 9th International Joint Conference on Natural Language Processing (EMNLP-IJCNLP)",
    month = nov,
    year = "2019",
    pages = "3982--3992"
}

@article{chen2025persona,
  title={Persona vectors: Monitoring and controlling character traits in language models},
  author={Chen, Runjin and Arditi, Andy and Sleight, Henry and Evans, Owain and Lindsey, Jack},
  journal={arXiv preprint arXiv:2507.21509},
  year={2025}
}

@inproceedings{kim2024prometheus,
    title = "Prometheus 2: An Open Source Language Model Specialized in Evaluating Other Language Models",
    author = "Kim, Seungone  and
      Suk, Juyoung  and
      Longpre, Shayne  and
      Lin, Bill Yuchen  and
      Shin, Jamin  and
      Welleck, Sean  and
      Neubig, Graham  and
      Lee, Moontae  and
      Lee, Kyungjae  and
      Seo, Minjoon",
    booktitle = "Proceedings of the 2024 Conference on Empirical Methods in Natural Language Processing",
    month = nov,
    year = "2024",
    pages = "4334--4353"
}

@article{wood1976role,
  title={The role of tutoring in problem solving},
  author={Wood, David and Bruner, Jerome S and Ross, Gail},
  journal={Journal of child psychology and psychiatry},
  volume={17},
  number={2},
  pages={89--100},
  year={1976},
  publisher={Blackwell Publishing Ltd Oxford, UK}
}

@article{rosenshine2012principles,
  title={Principles of instruction: Research-based strategies that all teachers should know.},
  author={Rosenshine, Barak},
  journal={American educator},
  volume={36},
  number={1},
  pages={12},
  year={2012},
  publisher={ERIC}
}

@article{graesser1995collaborative,
  title={Collaborative dialogue patterns in naturalistic one-to-one tutoring},
  author={Graesser, Arthur C and Person, Natalie K and Magliano, Joseph P},
  journal={Applied cognitive psychology},
  volume={9},
  number={6},
  pages={495--522},
  year={1995},
  publisher={Wiley Online Library}
}

@article{sweller1985use,
  title={The use of worked examples as a substitute for problem solving in learning algebra},
  author={Sweller, John and Cooper, Graham A},
  journal={Cognition and instruction},
  volume={2},
  number={1},
  pages={59--89},
  year={1985},
  publisher={Taylor \& Francis}
}

@article{shute2008focus,
  title={Focus on formative feedback},
  author={Shute, Valerie J},
  journal={Review of educational research},
  volume={78},
  number={1},
  pages={153--189},
  year={2008},
  publisher={Sage Publications}
}

@article{d2012dynamics,
  title={Dynamics of affective states during complex learning},
  author={D’Mello, Sidney and Graesser, Art},
  journal={Learning and Instruction},
  volume={22},
  number={2},
  pages={145--157},
  year={2012},
  publisher={Elsevier}
}

@inproceedings{liu2024personality,
  title={Personality-aware student simulation for conversational intelligent tutoring systems},
  author={Liu, Zhengyuan and Yin, Stella Xin and Lin, Geyu and Chen, Nancy},
  booktitle={Proceedings of the 2024 Conference on Empirical Methods in Natural Language Processing},
  pages={626--642},
  year={2024}
}

@article{mazare2018training,
  title={Training millions of personalized dialogue agents},
  author={Mazar{\'e}, Pierre-Emmanuel and Humeau, Samuel and Raison, Martin and Bordes, Antoine},
  journal={arXiv preprint arXiv:1809.01984},
  year={2018}
}

@article{serapio2023personality,
  title={Personality traits in large language models},
  author={Serapio-Garc{\'\i}a, Gregory and Safdari, Mustafa and Crepy, Cl{\'e}ment and Sun, Luning and Fitz, Stephen and Abdulhai, Marwa and Faust, Aleksandra and Matari{\'c}, Maja},
  year={2023}
}

@article{duan2026kaser,
  title={KASER: Knowledge-Aligned Student Error Simulator for Open-Ended Coding Tasks},
  author={Duan, Zhangqi and Fernandez, Nigel and Lan, Andrew},
  journal={arXiv preprint arXiv:2601.06633},
  year={2026}
}

@inproceedings{scarlatos2025exploring,
  title={Exploring knowledge tracing in tutor-student dialogues using llms},
  author={Scarlatos, Alexander and Baker, Ryan S and Lan, Andrew},
  booktitle={Proceedings of the 15th international learning analytics and knowledge conference},
  pages={249--259},
  year={2025}
}

@misc{huang2026interpretable,
  title         = {Interpretable Difficulty-Aware Knowledge Tracing in Tutor-Student Dialogues},
  author        = {Huang, Shuyan and Scarlatos, Alexander and Lee, Jaewook and Lan, Andrew},
  year          = {2026},
  eprint        = {2605.01097},
  archivePrefix = {arXiv},
  primaryClass  = {cs.CL},
  url           = {https://arxiv.org/abs/2605.01097}
}

\appendix
\section{Appendix}
\label{sec:appendix}
As mentioned earlier, our prompting and supervised fine-tuning (SFT) approach builds on the framework of \citet{scarlatos2026simulated}, using the code available on \url{https://github.com/umass-ml4ed/sim-student-eval}. In this section, we outline the setup required to train a steering vector.
\subsection{Implementation Details}

We implement the method in two stages: (i) construction of turn-level preference pairs from tutor-student dialogues, and (ii) optimization of a shared steering direction \(v\) together with tutor-specific scaling coefficients \(\{\delta_i\}_{i=1}^I\). The implementation follows the formulation in the main text, but includes several practical choices for efficient training and evaluation.


\paragraph{Sequence log-likelihoods.} Given a prompt-response pair \((q,r)\), we compute the conditional sequence log-likelihood \[ \log \pi_\theta(r \mid q) \] by summing next-token log-probabilities over the response tokens only. Prompt tokens are masked out and do not contribute to the score. These log-likelihoods are computed both for the unsteered model and for the steered model. Since the baseline values \[ \log \pi_\theta(t_{j,k}^i \mid q_{j,k}^i), \qquad \log \pi_\theta(\bar t_{j,k} \mid q_{j,k}^i) \] are reused throughout optimization, they are precomputed and cached for efficiency.

\paragraph{Parameterization of tutor-specific strength.}
As in the main text, we learn a shared steering direction \(v\) and tutor-specific coefficients \(\delta_i\). To enforce positivity and remove scale ambiguity, we optimize unconstrained parameters \(u_i \in \mathbb{R}\) and set
\[
\delta_i
=
\frac{\exp(u_i)}
{\frac{1}{I}\sum_{m=1}^{I}\exp(u_m)}.
\]
This ensures \(\delta_i > 0\) and \(\frac{1}{I}\sum_{i=1}^I \delta_i = 1\). In practice, we initialize \(u_i=0\) for all tutors, so that \(\delta_i=1\) initially, and initialize \(v\) from a Gaussian distribution. 

\paragraph{Training loss.}
While the main text presents the objective at the turn level, in implementation we group examples by \((j,i)\), i.e., all tutor turns from dialogue \(d_j^i\) associated with tutor \(i\). For each turn \(k\) in such a group, we compute the margin
\[
\begin{aligned}
m_{j,k}^i
&=
\bigl[
\log \pi_\theta(t_{j,k}^i \mid C_i)
-
\log \pi_\theta(\bar t_{j,k} \mid C_i)
\bigr]
\\
&\quad
-
\bigl[
\log \pi_\theta(t_{j,k}^i \mid C)
-
\log \pi_\theta(\bar t_{j,k} \mid C)
\bigr],
\end{aligned}
\]
where \(C_i=A_L(\cdot)+\delta_i v\) and \(C=A_L(\cdot)\).
The corresponding turn loss is
\[
\ell_{j,k}^i = -\log \sigma(\beta m_{j,k}^i).
\]
We then average \(\ell_{j,k}^i\) across turns \(k\) within a given \((j,i)\) group, and average again across groups in the minibatch. Thus, the implementation preserves the same BiPO-style preference structure as the main objective, while stabilizing optimization by aggregating at the dialogue-tutor level. 

\paragraph{Steering intervention.}
During training and inference, the intervention is applied by adding the scaled steering vector to the hidden activations at a designated transformer layer:
\[
A_L(\cdot)\leftarrow A_L(\cdot)+\alpha\,\delta_i v.
\]
During training, \(\alpha\) is absorbed into the learned parameterization, whereas at test time we expose a global scaling factor \(\alpha\) to control overall steering strength independently of the relative tutor-specific coefficients \(\delta_i\). 

\paragraph{Optimization and model selection.}
We optimize only the steering parameters \(\{v,u_1,\dots,u_I\}\), keeping the underlying language model fixed. Validation loss is computed using the same grouped objective, with cached baseline log-likelihoods for the unsteered model. The best checkpoint is selected according to validation performance, and the learned parameters are then used for test-time generation.

\paragraph{Inference.}
At test time, for a target tutor \(i\), generation is performed with the intervention \(A_L(\cdot)+\alpha\,\delta_i v\). This yields tutor-conditioned outputs while retaining a single shared steering direction across all tutors. The unsteered SFT model serves as the baseline condition.

\begin{algorithm*}[h!]
\small
\caption{Learning a steering vector and activating steering using the vector}
\label{alg:our_impl}
\DontPrintSemicolon
\KwIn{Dialogue dataset \(\mathcal{D}\), SFT model \(\pi_\theta\), steering layer \(L\), temperature \(\beta\), training steps \(T\)}
\KwOut{Shared steering vector \(v\), tutor-specific coefficients \(\{\delta_i\}_{i=1}^I\)}

\textbf{Pair construction:} \;
\ForEach{tutor \(i\), dialogue \(d_j^i\), tutor turn \(k\)}{
Construct prompt \(q_{j,k}^i\) from the dialogue context\;
Set chosen response to \(t_{j,k}^i\)\;
Generate rejected response \(\bar t_{j,k} \sim \pi_\theta(\cdot \mid q_{j,k}^i)\)\;
Store pair \(\bigl(q_{j,k}^i, t_{j,k}^i, \bar t_{j,k}\bigr)\)\;
}
Precompute and cache
\(\log \pi_\theta(t_{j,k}^i \mid q_{j,k}^i)\) and
\(\log \pi_\theta(\bar t_{j,k} \mid q_{j,k}^i)\)\;

\BlankLine
\textbf{Initialization:} \;
Initialize \(u_i \leftarrow 0\) for all \(i\), and initialize \(v\) randomly\;
Set
\[
\delta_i \leftarrow \frac{\exp(u_i)}{\frac{1}{I}\sum_{m=1}^I \exp(u_m)}.
\]
\textbf{Training:} \;
\For{\(t = 1, \dots, T\)}{
Group examples by \((j,i)\)\;
\ForEach{group \((j,i)\)}{
\ForEach{turn \(k\) in \(d_j^i\)}{
Compute steered log-likelihoods
\[
\log \pi_\theta\!\bigl(t_{j,k}^i \mid A_L(\cdot)+\delta_i v\bigr),
\quad
\log \pi_\theta\!\bigl(\bar t_{j,k} \mid A_L(\cdot)+\delta_i v\bigr)
\]
and retrieve cached baseline log-likelihoods
\[
\log \pi_\theta\!\bigl(t_{j,k}^i \mid A_L(\cdot)\bigr),
\quad
\log \pi_\theta\!\bigl(\bar t_{j,k} \mid A_L(\cdot)\bigr).
\]
Form margin
\[
m_{j,k}^i =
\Bigl[\log \pi_\theta(t_{j,k}^i \mid A_L+\delta_i v)
-\log \pi_\theta(\bar t_{j,k} \mid A_L+\delta_i v)\Bigr]
-
\Bigl[\log \pi_\theta(t_{j,k}^i \mid A_L)
-\log \pi_\theta(\bar t_{j,k} \mid A_L)\Bigr].
\]
Compute turn loss
\[
\ell_{j,k}^i = -\log \sigma(\beta m_{j,k}^i).
\]
}
Average \(\ell_{j,k}^i\) over turns \(k\) in group \((j,i)\)\;
}
Average loss over groups in the minibatch and update \(v\) and \(\{u_i\}\)\;
Recompute \(\delta_i \leftarrow \exp(u_i)/(\frac{1}{I}\sum_m \exp(u_m))\)\;
}

\BlankLine
\textbf{Inference:} \;
\For{target tutor \(i\)}{
Apply activation steering
\[
A_L(\cdot) \leftarrow A_L(\cdot) + \alpha\,\delta_i v
\]
during autoregressive generation\;
}
\end{algorithm*}

\begin{table*}[t]
\small
\centering
\caption{Examples of our DPO dataset to train a steering vector (Part 1).}
\label{tab:appendix_drastic_diff_examples_1}
\begin{tabular}{p{0.05\textwidth} p{0.45\textwidth} p{0.22\textwidth} p{0.22\textwidth}}
\toprule
\textbf{ID} & \textbf{Prompt} & \textbf{Chosen} & \textbf{Rejected} \\
\midrule

12024
&
[system]\newline
Cutting Knowledge Date: December 2023\newline
Today Date: 26 Jul 2024\newline
\newline
You are a tutor guiding a student through a math problem.\newline
\newline
[user]\newline
Question: Work out the area of the shape [Image: A composite shape made up of a rectangle and a triangle with some side measurements labelled]\newline
A. $68 \mathrm{~m}^{2}$\newline
B. $34 \mathrm{~m}^{2}$\newline
C. $36 \mathrm{~m}^{2}$\newline
D. $44 \mathrm{~m}^{2}$\newline
\newline
(No First Student Turn)\newline
\newline
[assistant]\newline
Hi! How can I help?\newline
\newline
[user]\newline
I dont really understand this question can you please help me with it\newline
\newline
[assistant]\newline
Of course, can you tell me how you started? Or were you just not sure at all?\newline
\newline
[user]\newline
im not sure how i started it\newline
\newline
[assistant]\newline
Ok, well I would start by splitting up the shape into a rectangle and a triangle\newline
\newline
[user]\newline
ok\newline
\newline
[assistant]\newline
Can you tell me the area of that rectangle on the left?\newline
\newline
[user]\newline
24m squared\newline
\newline
[assistant]
&
Fab. Now for the triangle, the sides aren't labelled yet. So if the whole width of the shape is 10m, and the rectangle is 4m, could you work out how wide the triangle is?
&
Great!
\\

\bottomrule
\end{tabular}
\end{table*}

\begin{table*}[t]
\small
\centering
\caption{Examples of our DPO dataset to train a steering vector (Part 2).}
\label{tab:appendix_drastic_diff_examples_2}
\begin{tabular}{p{0.05\textwidth} p{0.45\textwidth} p{0.22\textwidth} p{0.22\textwidth}}
\toprule
\textbf{ID} & \textbf{Prompt} & \textbf{Chosen} & \textbf{Rejected} \\
\midrule

1988
&
[system]\newline
Cutting Knowledge Date: December 2023\newline
Today Date: 26 Jul 2024\newline
\newline
You are a tutor guiding a student through a math problem.\newline
\newline
[user]\newline
Question: $\frac{1}{2} \times 3=$\newline
A. $\frac{3}{2}$\newline
B. $3 \frac{1}{2}$\newline
C. $\frac{3}{6}$\newline
D. $\frac{1}{6}$\newline
\newline
First Student Turn: hi\newline
\newline
[assistant]\newline
hello \raisebox{-0.2em}{\includegraphics[height=1em]{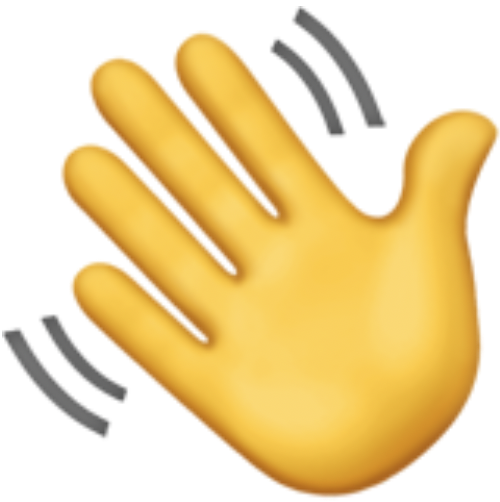}}  how can I help you?\newline
\newline
[user]\newline
all of this\newline
\newline
[assistant]\newline
I'm just reading the question let's think about 3 lots of 1/2 so first I'm going to think about what I get if I add 3 of them together what is 1/2 + 1/2?\newline
\newline
[user]\newline
1\newline
\newline
[assistant]\newline
great! and then add another 1/2. so what is 1 + 1/2?\newline
\newline
[user]\newline
1 and 1/2\newline
\newline
[assistant]\newline
\raisebox{-0.1em}{\includegraphics[height=1em]{thumbs_up.png}}  \raisebox{-0.2em}{\includegraphics[height=1em]{smile.png}}  so the answer to three times 1/2 must be 1 and 1/2 oh, look at the options in the question. which is the same as 1 and 1/2\newline
\newline
[user]\newline
1/6\newline
\newline
[assistant]
&
no sorry, 1/6 is much smaller than 1/2 it's like you share something between 6 people instead of just 2 people. Each person gets less when there's 6 people to share something between instead of just 2 people sharing it have another look at the ABCD options we now know it's not D
&
A is 3/2 \raisebox{-0.2em}{\includegraphics[height=1em]{smile.png}} \raisebox{-0.2em}{\includegraphics[height=1em]{smile.png}} \raisebox{-0.2em}{\includegraphics[height=1em]{smile.png}} 
\\

\bottomrule
\end{tabular}
\end{table*}

\end{document}